This paper has been accepted for publication at the IEEE Integrated STEM Education Conference. Please cite the paper as: L. Carlone, K. Khosoussi, V. Tzoumas, G. Habibi, M. Ryll, R. Talak, J. Shi, and P. Antonante. Visual navigation for autonomous vehicles: An open-source hands-on robotics course at MIT. In IEEE Integrated STEM Education Conference (ISEC), 2022.# Visual Navigation for Autonomous Vehicles: An Open-source Hands-on Robotics Course at MIT

Luca Carlone
Massachusetts Institute of Technology, lcarlone@mit.edu

Kasra Khosoussi, Vasileios Tzoumas, Golnaz Habibi, Markus Ryll,
Kasra.Khosoussi@data61.csiro.au, vtzoumas@umich.edu, golnaz.habibi@gmail.com, markus.ryll@tum.de

Rajat Talak, Jingnan Shi, Pasquale Antonante
Massachusetts Institute of Technology, talak, jnshi, antonap@mit.edu*Abstract* - This paper reports on the development, execution, and open-sourcing of a new robotics course at MIT. The course is a modern take on "Visual Navigation for Autonomous Vehicles" (VNAV) and targets first-year graduate students and senior undergraduates with prior exposure to robotics. VNAV has the goal of preparing the students to perform research in robotics and vision-based navigation, with emphasis on drones and self-driving cars. The course spans the entire autonomous navigation pipeline; as such, it covers a broad set of topics, including geometric control and trajectory optimization, 2D and 3D computer vision, visual and visual-inertial odometry, place recognition, simultaneous localization and mapping, and geometric deep learning for perception. VNAV has three key features. First, it bridges traditional computer vision and robotics courses by exposing the challenges that are specific to embodied intelligence, e.g., limited computation and need for just-in-time and robust perception to close the loop over control and decision making. Second, it strikes a balance between depth and breadth by combining rigorous technical notes (including topics that are less explored in typical robotics courses, e.g., on-manifold optimization) with slides and videos showcasing the latest research results. Third, it provides a compelling approach to hands-on robotics education by leveraging a physical drone platform (mostly suitable for small residential courses) and a photo-realistic Unity-based simulator (open-source and scalable to large online courses). VNAV has been offered at MIT in the Falls of 2018-2021 and is now publicly available on MIT OpenCourseWare (OCW) and at *vnav.mit.edu/*.

*Index Terms* – Open-source Course Material, Project-based Learning, Robotics and Computer Vision Education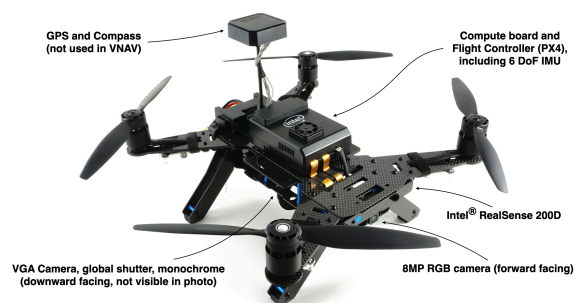

(a) Drone Platform

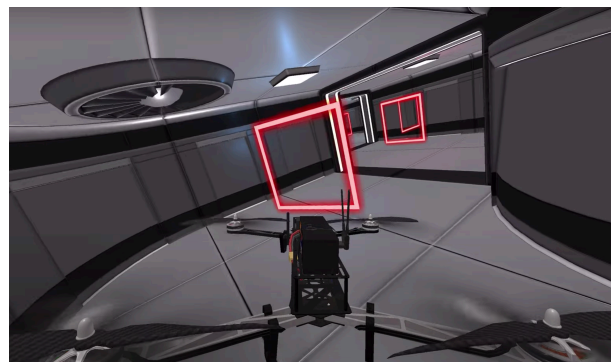

(b) Unity-based Simulator

FIGURE I
HARDWARE PLATFORM & SIMULATION INFRASTRUCTURE USED IN VNAV

## INTRODUCTION

Robotics and autonomous systems are steadily transforming society by revolutionizing transportation, manufacturing, supply chain logistics, aviation, disaster response and national security, among other domains. This fast-paced transformation puts an increasing and urgent emphasis on training the next generation of engineers and scientists that will be in charge of conceiving, designing, implementing, and operating these systems.

**Challenges and opportunities of robotics education.** Robotics education poses novel challenges to educators. First, robotics education cuts across multiple disciplines (e.g., computer science, electrical/mechanical/aerospace engineering, applied mathematics) and research areas (e.g., computer vision, machine learning, control theory), hence any single-semester course has to find a suitable trade-off between depth and breadth, as well as between theory and applications. Second, robotics research and applications are advancing at an ever-increasing pace, hence challenging educators to continuously adjust the learning objectives to keep up with the changing demands of a flourishing job market. For instance, the last decade has simultaneously witnessed a boom in self-driving applications and a radical paradigm shift where model-based approaches have been challenged by learning-based data-driven methods. Third, advanced robotics applications often require multiple sensors (e.g., 3D lidars, commercial RGB-D cameras) and powerful embedded computers that are often too expensive to purchase and maintain for residential project-based education and become inadequate for online offerings.

At the same time, these challenges offer unprecedented opportunities for students and educators. First, the fact that robotics spans multiple communities and disciplines makes it a natural playground to exercise foundational knowledge (e.g., linear algebra, optimization, control) while reinforcing it through deliberate practice (e.g., implementing and testing algorithms on a robot). Second, the presence of a vibrant research community and the growing interest towards autonomous systems from the media provide a natural motivation for students, by exposing a clear set of stakeholders that can use the expertise developed in robotics courses. Finally, one realizes that the need for sensors and embedded computing is a "feature" rather than a "bug": robotics is the science of embodied intelligence and the goal of processing information in real-time subject to limited computation is an essential ingredient rather than an afterthought. In addition to this pedagogical opportunity, concerns about the cost of sensors and computation have been partially alleviated by two trends: (i) advances in chip and sensor design –triggered by their use in commercial products, such as smart phones– have led to a new generation of sensors and computers whose cost and form-factor become increasingly compatible with residential education; (ii) advances in simulation have led to the development of lightweight and photo-realistic simulators, which are now commonly used in industry and academia to exercise robotics algorithms and implementations.

**The untapped potential of visual navigation.** Within the broad landscape of robotics education, visual navigation (where "visual" is meant in a broad sense to include information-rich onboard sensors such as cameras and 3D lidars) has quickly become a fundamental area of expertise. Visual navigation powers modern autonomous vehicles, from self-driving cars, to drones for disaster response, last-mile delivery, and precision agriculture. However, few courses are currently capturing the complexity and untapped educational opportunities in this area. Expertise contributing to visual navigation is fragmented across different courses focusing on robotics and control, computer vision, and machine learning. This fragmentation has several downsides: (i) traditional robotics and control courses often prefer focusing on simpler sensors (e.g., 2D lidars or ultrasonic sensors), hence creating a disconnect with many modern robotics applications; (ii) computer vision courses typically do not "close the loop" over control and actuation, hence de-emphasizing the importance of just-in-time computation; and (iii) machine learning courses typically do not focus on robotics applications, hence missing an opportunity to stress the implications of learning-induced failures on safety critical applications and also creating a disconnect with traditional model-based approaches in robotics and vision.

**Contribution.** In this paper, we attempt to bridge this gap by reporting on the development, execution, and open-sourcing of a new robotics course at MIT. The course is a modern take on "Visual Navigation for Autonomous Vehicles" (VNAV) and targets first-year graduate students and senior undergraduates with prior exposure to robotics. VNAV aims at preparing students to perform research in robotics and vision-based navigation (or, more generally, to design, develop, and operate advanced robotics systems), with emphasis on drones and self-driving cars. The course covers a broad set of topics, including geometric control and trajectory optimization, 2D and 3D computer vision, visual and visual-inertial odometry, place recognition, simultaneous localization and mapping, and geometric deep learning for perception. VNAV has three key features:

- it bridges traditional computer vision and robotics courses by exposing the challenges that are specific to embodied intelligence, e.g., limited computation and need for just-in-time processing and robust perception to close the loop over control and decision making. This is done by combining selected topics in control with advanced topics in robot vision and perception;
- it strikes a balance between depth and breadth by combining rigorous technical notes with slides and videos showcasing recent research advances. For instance, the course starts with a basic introduction to 3D geometry, using a Lie-group theoretic lens, touches on advanced topics, such as optimization on manifolds, and stretches to the latest advances in geometric deep learning and graph neural networks;
- it provides a compelling approach for hands-on robotics education by leveraging a physical drone platform (mostly suitable for small residential courses) and a photo-realistic Unity-based simulator (open-source and scalable to large online courses).

In the rest of this paper, we discuss the VNAV curriculum, software and hardware infrastructure, and we report on students' feedback and lessons learned. VNAV has been offered at MIT in 2018-2021 and is now publicly available on MIT OpenCourseWare (OCW) [28] and at *vnav.mit.edu/*. Moreover, courses based on VNAV have been also offered at the University of Michigan and at the Technical University of Munich in 2021.

### RELATED COURSES

Robotics education is a well-studied topic. An IEEE Xplore search for the keywords "robotics & education & course" shows more than 650 publications after the year 2000.

The existing literature has extensively stressed the importance of hands-on labs [1]-[3]. Moreover, the use of final projects and competitions has been understood as a powerful approach to improve knowledge assimilation, and to stress that there can be more than one correct answer to a problem [4]-[6]. Researchers and educators have also investigated flipped classroom approaches [7], where the classroom time has a stronger focus on hands-on activities. Several courses are at the introductory level and target first-year undergraduate students [8] or high school students [9], where the goal is retention rather than preparing the students for research, development, and operation of advanced robotics systems. Other educators have proposed multi-semester courses to deal with the breadth of robotics topics to cover [10]. Traditional robotics courses focus on control algorithms and use simple sensors and actuators (e.g., LEGO Mindstorm [11]). Efforts to increase scalability and reduce costs have focused on the development of simulators [12] or web-based technologies to enhance learning [13], which however have mostly targeted robotics manipulation.

The literature on visual navigation education is sparser and more recent. If we also add the keyword "computer vision" to the IEEE Xplore search mentioned above, then the number of results drops to 36. Few courses focus on advanced vision-based navigation due to the challenges mentioned in the introduction. Early attempts to infuse visual-navigation in mobile robotics [14] trace back to before the "deep learning revolution" and before most of the latest advanced in visual localization and mapping. Maxwell and Meeden [14] focus on early vision (e.g., edge detection, stereo depth). Nitschke et al. [15] propose a one-day contest on automatic visual drone navigation, but mostly focuses on control aspects. Paull et al. [16] propose "Duckietown", a course meant to exercise vision-based navigation for self-driving cars; the course has a broader scope and while it has the advantage of leveraging relatively inexpensive robots (e.g., a differential drive wheeled platform equipped with a Raspberry PI and a camera), these robots do not have enough computation to exercise advanced vision-based localization and mapping algorithms. Brand et al. [17] and Eller et al. [18] develop the "PiDrone" platform, an exciting low-cost aerial platform for robotics education. The autonomy stack is developed in Python and uses the Robot Operating System (ROS). Other efforts have focused on middle school and high school outreach, with emphasis on mobile robotics [2], [3] and manipulation [19].

Finally, it is worth reviewing relevant online courses and material. Thrun's seminal course on "Artificial Intelligence for Robotics" [20] on Udacity covers selected topics in planning, localization, tracking and control using self-driving cars as the main motivation. Daniilidis and Shi offer the course "Robotics: Perception" [21] on Coursera; the course covers from basic geometric and computer vision to multi-view geometry. Waslander and Kelly offer a "Self-Driving Cars Specialization" [22] on Coursera which leverages an advanced driving simulator. Roy et al. [47] offer the course "Flying Car and Autonomous Flight Engineer" on Udacity, which focuses on control, planning, and GPS-based state estimation. Stachniss [48] focuses on mapping for robotics, with full lectures and short online videos. Recent courses on edX share similar motivations as ours but differ in the selection of topics [23], [24]. It is also worth mentioning broader open-source initiatives, such as the F1Tenth initiative [25] and BWSI (Beaver Works Summer Institute) [26], that provide introductory-level courses for seniors and K-12 students.

## THE VNAV CURRICULUM

### I. Prerequisites, Topics, and Learning Goals

VNAV assumes basic familiarity with C++ programming (e.g., syntax, function calls, compilation and execution of programs using *CMake*), linear algebra (e.g., matrix operations, eigenvalues/eigenvectors, matrix factorization), and control theory (e.g., dynamical systems, PID control).

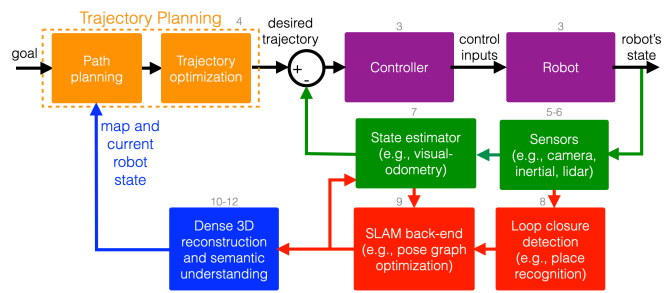

FIGURE II
CANONICAL ARCHITECTURE USED IN VNAV

The course covers the building blocks and the entire architecture of a modern vision-based autonomous navigation system, as shown in Figure II. The system receives as input a target location the robot has to reach and plans a path to that location (orange blocks). The path is then followed using a suitable controller to govern the robot dynamics (purple blocks). In vision-based navigation, the state of the system and the external environment are not known a priori, and must be estimated from sensor data. The blocks in green represent sensing and robot state estimation algorithms (typically executed at high-rate to provide state feedback to the controller). The blocks in red and blue instead lead to the estimation of the map of the environment the robot is moving in, which is used to inform planning algorithms (typically, both planning and mapping are executed at a lower rate). VNAV covers all these modules (the number above each module in Figure II corresponds to a week in the schedule in Table I): while the perception modules (blocks in red, green, blue) are presented in full generality and are applicable to common sensing modalities (e.g., cameras, lidars, IMU), the planning and control blocks are tailored to agile drone navigation, which –contrarily to wheeled platforms– allow highlighting the complexity of navigation in 3D environments. The VNAV labs focus on implementing each module in the architecture, and gradually build towards an implementation of the entire visual-navigation pipeline in Figure II.

TABLE I
COURSE STRUCTURE AND TOPICS

| Week | Lecture Topics | Lab Topics |
|---|---|---|
| 1 | Introduction | Lab 1: Linux, C++, Git |
| 2 | 3D Geometry and Lie Groups | Lab 2: ROS |
| 3 | Geometric Control | Lab 3: 3D trajectory following |
| 4 | Trajectory Optimization | Lab 4: 3D trajectory optimization |
| 5 | 2D Computer Vision | Lab 5: Feature detection |
| 6 | 2-view Geometry and RANSAC | Lab 6: Feature tracking and matching |
| 7 | Multi-View Geometry, Visual Odometry, and Optimization | Lab 7: GTSAM |
| 8 | Place Recognition | Lab 8: ML for robotics |
| 9 | SLAM and Visual-Inertial Navigation | Lab 9: SLAM |
| 10 | Active Research Areas in Robot Perception | Final project |
| 11 | Robust Estimation and Dense 3D Reconstruction | Final project |
| 12 | Semantic Understanding and Geometric Deep learning | Final project |
| 13 | Guest Lectures | Final project |
| 14 | Students Presentations | |

The learning goals for the VNAV students consist in developing the ability to:

- **recall and understand** basic mathematical notions in numerical linear algebra, Lie groups, and optimization on manifolds, and their application to visual navigation;
- **solve novel problem instances** (from computer vision to trajectory optimization problems) using proof techniques seen in class, and formally **present** their derivation and results using proper terminology;
- **read, understand, and critically analyze** technical papers and test state-of-the-art algorithms available in standard open-source libraries;
- critically analyze the performance of given software implementations, **perform rigorous experimental evaluation**, and discuss the impact of different algorithmic and parameter choices;
- **improve** existing methods, implement them on a real or simulated platform, and demonstrate their effectiveness.

*II. Course Structure*

The learning goals are achieved in a 14-week schedule (Table II), which alternates frontal lectures with hands-on labs, and concludes with an open-ended final project. We describe each component of the VNAV schedule below.

Lectures are offered three times a week; each lecture is designed to be 50 minutes long. The lectures are mostly done on the board and cover the mathematical foundations, but also alternate mathematical concepts on the board with slides (typically presented at the beginning or the end of lecture) to provide motivations or show applications of the theoretical concepts. The lectures are supported by lecture notes (also released at *vnav.mit.edu/*). Table I shows the topics for each week of lecture; the table is color-coded using the same colors used to describe the blocks in Figure II. The color code highlights how the lectures cover the entire architecture, while emphasizing aspects related to perception and vision. The first two weeks are used to refresh basic concepts in linear algebra and 3D geometry, as well as provide a pragmatic introduction to manifolds and Lie groups. Weeks 3 to 12 cover the core material of the course, including geometric control, trajectory optimization, 2D and 3D computer vision (e.g., feature detection and matching, 2-view and multi-view geometry, RANSAC), optimization on manifolds, visual and visual-inertial odometry, place recognition, and Simultaneous Localization and Mapping (SLAM). These weeks conclude with advanced topics, including robust estimation, dense 3D reconstruction, and geometric deep learning for semantic understanding. The last two weeks of VNAV are devoted to guest lectures and students' presentations. Guest lecturers from industry are invited to discuss other advanced topics or to highlight applications of the VNAV material to real problems; for instance, we invited speakers from Skydio to showcase vision-based drone navigation, or speakers from Boston Dynamics to discuss additional challenges arising from legged locomotion (not covered in VNAV).

Labs complement the theoretical foundations developed in the lectures. VNAV includes 2-hour-long weekly labs with hands-on activities. The first two labs are individual and introduce key software tools used in VNAV (more details in the "Hardware and Software Infrastructure" section below), including Git [29] (for version control) and ROS (the Robot Operating System) [30]. Moreover, these introductory labs provide a refresher about C++ and Linux. Starting from Lab 3, students form teams of 2 or 3, and work together towards implementing instances of the autonomy blocks in Figure II (we refer the reader to the color code used in Figure II and Table I, that describes the correspondence between labs and autonomy blocks). Contrarily to introductory courses, the labs are designed to exercise advanced open-source libraries (e.g., OpenCV [31], GTSAM [32], ORB-SLAM [33-35], DBoW2 [36]) rather than implementing capabilities from scratch, which more closely mimics the typical robotics research and development process. Moreover, the lab exercises stress potential failure cases of existing algorithms and the importance of designing trustworthy algorithms and implementations to support safety critical applications of autonomous vehicles. The labs rely on the platforms discussed in the "Hardware and Software Infrastructure" section below and alternate more abstract coding exercises with deployment and testing on the robot and simulator. For instance, Lab 3 includes a simpler control exercise based on our Unity simulator, which aims at developing a geometric control scheme to have the drone fly as quickly as possible along a circular pattern. On the other hand, Lab 4 considers a more realistic and compelling application scenario where the students have to design a trajectory optimization scheme for drone racing, where the drone has to traverse a sequence of gates in the shortest time (see Figure I(b) and the following short video *youtu.be/ssgfN7l4STI* for an example). Labs 5-9 introduce the students to four open-source libraries, OpenCV [31], GTSAM [32], ORB-SLAM [33-35], DBoW2 [36], which are broadly used by researchers and practitioners.

The course concludes with a 4-week-long final project. The instructors provide final project ideas, grouped in three types of projects:

- **survey projects** do not directly require implementation but rather consist in reviewing and presenting technical papers on topics that were not cover in the lectures and labs; in particular, these projects are used to cover more advanced topics in robotics research (e.g., novel applications of deep learning to SLAM, localization in challenging visual conditions);
- **system projects** consist in replicating an existing result in the literature; these projects involve a substantial amount of implementation and testing and may involve the use of robotics platforms or standard benchmarking datasets; these projects are often a good starting point for students approaching robotics research and exploring potential research opportunities. Moreover, they align with existing international initiatives, such as the *ML reproducibility challenge* [37];
- **research projects** target students who have already started doing robotics research and may benefit from making progress and getting early/external feedback on their work. These projects are often student-driven, but may also arise from project ideas proposed by the instructors.

The final project ideas are typically inspired by real-world use cases. For instance, multiple research projects (e.g., mapping the MIT tunnels, multi-sensor calibration) have been inspired by the DARPA Subterranean Challenge [43], a research-oriented competition aimed at developing autonomous systems capable of mapping underground environments and reporting the location of elements of interest (e.g., survivors).

The VNAV structure is largely inspired by the CDIO (Conceive Design Implement Operate) approach [38], in that it features extensive group projects and project-based learning, and provides opportunities for the students to exercise technical knowledge as well as communication skills. The final projects lead to a final report and presentation, both designed to roughly mimic the publication and conference presentation process. The report is done using a standard robotics paper format and the evaluation is similar to a peer review process in robotics. The final projects are then presented to the rest of the students in a 20-minute frontal presentation. The presentations for system and research projects typically include videos and demo showcasing the robotics systems and algorithms developed during the final projects.

*II. Resources and Grading*

While the course is mostly self-contained, two main textbooks are also suggested as a more in-depth complement to the lecture notes. In particular, VNAV adopts Barfoot's "State Estimation for Robotics" book [27] and Yi et al.'s "An Invitation to 3-D Vision: From Images to Geometric Models" book [39]. Some portions of VNAV are inspired by Dellaert and Kaess' monograph "Factor Graph for Robot Perception" [40]. Finally, each lecture also provides pointers to relevant references (e.g., tutorial/survey papers as well as technical papers), which target specific approaches (e.g., polynomial trajectory optimization) or more recent topics that are still subject of active research (e.g., graph neural networks for robot perception).

The students are graded based on the lab exercises and the final project. The lab exercises (also released at *vnav.mit.edu/*) include a set of questions covering both theoretical aspects (discussed during the lectures) and the result of implementation and testing of algorithms (assigned during the labs). Initial assignments (Lab 1-2) are individual, while Labs 3-9 include a mix of team assignments and individual assignments. The lab grades account for 65% of the final grade, while the final project accounts for 25%. Both labs and final project handouts outline a clear rubric for grading. The remaining 10% of the grade is based on TA and team-members' evaluations: each student is assessed by their teammates and the TAs in order to more fairly assess their contribution to their team's assignments.

### HARDWARE AND SOFTWARE INFRASTRUCTURE

This section describes the hardware platform and the simulator used in VNAV, as well as the software infrastructure developed to support the course.

*I. Hardware Platforms and Simulator*

During the first two years, VNAV used a physical drone to support the labs and the final projects. The platform has been replaced by a photo-realistic simulator after the COVID-19 outbreak in 2020. We describe both below.

The hardware platform was based on the *Intel Aero Ready To Fly* quadrotor [41] –see Figure I(a)– which is customizable and can be easily interfaced with the *Robot Operating System* (ROS) and standard motion capture systems (e.g., OptiTrack or Vicon). The quadrotor is equipped with an embedded real-time flight controller running the PX4 Autopilot [42] combined with a more capable computer (the Intel Aero Compute Board) dedicated to running ROS. The quadrotor is equipped with an Inertial Measurement Unit (IMU), and a RealSense depth camera. The drone, which has been recently discontinued, costed around $2,500. We are currently working on developing a custom platform, with similar sensing and actuation, but using a more powerful NVIDIA Xavier NX [44] computer. An interesting alternative is the PiDrone platform [18].

After the COVID-19 outbreak, VNAV's labs have transitioned to use a photo-realistic Unity-based simulator, named *TESSE*, which has been developed in collaboration with MIT Lincoln Laboratory and has been also released open-source at *github.com/MIT-TESSE*. In the context of VNAV, we have adapted TESSE to model realistic drone dynamics, simulate sensor data found on typical quadrotors (i.e., IMU and depth cameras), and re-create challenging testing scenarios based on a drone racing application. Figure I(b) reports a snapshot from the simulator, showcasing the drone and the gates (in red) used for drone racing. The simulator has enabled a virtual offering of VNAV during the COVID-19 pandemic and may be also used in future (larger-scale) online offerings of the course.

## II. Software Infrastructure

The VNAV software infrastructure relies on GNU/Linux as an operating system (Ubuntu distribution), and uses ROS (the Robot Operating System) [30] as a middleware. Version control and software distribution is based on Git [29]. These tools are commonly adopted in robotics research and development. All coding assignments are based on C++.

Within this environment, we make use of popular open-source libraries to support the different VNAV modules. The 2D computer vision algorithms rely on OpenCV [31]. The 3D geometric computer vision algorithms (from RANSAC to on-manifold optimization) rely on OpenGV [45] and GTSAM [32]. The trajectory optimization module relies on an open-source library from ETH [46]. The place recognition lab uses DBoW2 [36], which implements a bag-of-visual-words approach for place recognition. Finally, the SLAM lab relies on ORB-SLAM3 [35] as a library for visual and visual-inertial localization and mapping.

## ASSESSMENT AND DISCUSSION

This section provides an assessment of VNAV's learning outcomes and discusses potential improvements.

## I. Quantitative Assessment

Figure III reports aggregate statistics obtained from the student responses to the end-of-semester course evaluations. Figure III focuses on the evaluations measuring VNAV's impact in increasing the students' (i) interest on the subject (blue bars), (ii) knowledge spectrum (magenta bars), and (iii) ability to perform research (orange and yellow bars). Data are cumulative across all the course's offerings at MIT and the University of Michigan. The MIT data spans the period 2018-2021 (4 offerings), including 60 student evaluations (average response rate was 70%); the University of Michigan data was collected in 2021 (1 offering), including 36 student evaluations (response rate was 85%). The students responded with either Strongly Agree (SA), Agree (A), Neutral (N), Disagree (D), or Strongly Disagree (D). We observe that across all 4 subject questions (A to D above), at least 87-92% of the students responded with SA or A, at most 1-4% of the students responded with D or SD, and the rest 3-8% of the students responded with N.

Several observations are in order. First, VNAV increased the students' interest on the topic of vision-based autonomous navigation (cf. blue bars in Figure III). The students quoted in their written evaluations their excitement for being introduced to a whole new field. We believe that the dual focus of the course on theory and practice was one of the major reasons that helped increase the students' interest in the subject; we also believe that connecting labs and final projects to real-world applications, ranging from the DARPA Subterranean Challenge [43], to Mars rovers and autonomous drones' applications has contributed to increasing the students' interest.

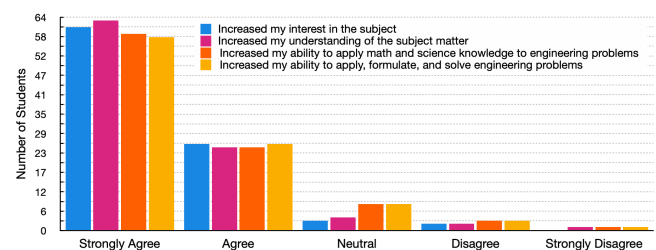

FIGURE III
END-OF-SEMESTER COURSE EVALUATIONS: SUMMARY STATISTICS

Second, VNAV largely increased the students' knowledge spectrum on autonomous navigation (cf. magenta bars in Figure III). We believe that the course's focus on teaching the entire visual-navigation pipeline allowed the students to acquire a holistic perspective of how real-world robots can perceive and navigate their surroundings. The students frequently quoted the valuable experience of both working on theory exercises and, at the same time, implementing software to enable real-time perception capabilities in real-world navigation scenarios.

Third, VNAV increased the students' ability to apply mathematical and scientific knowledge, formulate, and solve complex engineering problems (cf. yellow and orange bars in Figure III), which are key prerequisites for research. This observation is also supported by the fact that in the first 3 offerings of VNAV (from 2018 to 2020), students' final projects led to 12 peer-reviewed publications in top robotics conferences (such as ICRA, IROS, ACC, and ISER) and journals (IEEE Transactions on Robotics and IEEE Robotics and Automation Letters). VNAV students also developed and released two major open-source software libraries. The anonymous feedback provided in the subject evaluations suggests that VNAV has been successful in assisting young robotics researchers to further hone their research skills and ideas. That being said, additional data and further analysis are needed to formally establish a causal link between our approach in VNAV and students' research results, which is clearly impacted by factors external to VNAV and is biased by the fact that the student sample already includes research-inclined students with interest towards robotics.

## II. Potential Improvements

**Improving the software and hardware infrastructure**. The VNAV software infrastructure has been developed and refined over the 4 years the course has been offered at MIT. While the feedback from the students has been increasingly positive about the labs, it might be desirable to streamline the software installation and setup process. For instance, some students might spend time trying to compile specific pieces of software on their machines and, while we believe this is an important experience towards robotics research, it might detract from the time spent towards the key learning goals. A potential solution is to use Docker [49] to improve software portability across platforms. Regarding the hardware infrastructure, it would be desirable to develop cheaper drone platforms to broaden access to the course. An interesting effort in this direction is the PiDrone [18].

**Undergraduate offerings and outreach**. Currently, VNAV targets first-year graduate students and senior undergraduates with prior exposure to robotics. However, it would be interesting to broaden participation and adapt the course to be offered to undergraduates and high-school students with no robotics experience. In future offerings, we plan to open the MIT offering of VNAV to undergraduate students with background in control theory, linear algebra, optimization, and programming. We already had good success with the seniors attending VNAV and we believe that splitting the labs into baseline goals (targeting undergraduates) and advanced goals (targeting graduate students) might be an effective way to be more inclusive towards undergraduate students. Redesigning a selected portion of the course to support high-school outreach activities is more challenging but feasible. For instance, BWSI [26] has offered courses involving visual-navigation with drones, using a platform (similar to the one in Figure I(a)) designed in collaboration with the VNAV staff in 2018.

**Creating an open-source VNAV community**. VNAV has been offered in multiple universities, including MIT, the University of Michigan, and TUM. In the future, it would be desirable to establish bridges across the different offerings. A potential starting point is to create a joint repository of final projects, to increase the options the students can choose from and to avoid duplicating efforts in developing the project ideas and background material. For instance, the 2021 offering at the University of Michigan included a novel project on perception-aware Model Predictive Control (including description and background material) that might be useful for future offerings at other universities

**Visualization, visualization, visualization**. In past VNAV offerings, we used GeoGebra [50] to visualize key ideas when discussing optimization on manifolds. Students were able to interact with our tool through the GeoGebra web interface. This tool allowed students to visualize and interact with abstract mathematical concepts such as tangent spaces, tangent vectors, retractions, and geodesics. Additionally, we also used *g2o* [51] (and its GUI) in the classroom to visualize the behavior of various optimization methods for solving SLAM. The feedback provided by the students indicate that these visualization tools helped them grasp and internalize key concepts quickly, accurately, and intuitively. In future offerings of VNAV, we plan to further explore mathematical visualization, interactive learning, and live demonstrations using tools such as GeoGebra [50], Manim [52], and g2o [51].

## CONCLUSION

This paper reported on the development, execution, and open-sourcing of the course "Visual Navigation for Autonomous Vehicles" (VNAV) offered at MIT in the Falls of 2018-2021, which has also been adopted (with modifications) in the curricula at the University of Michigan and the Technical University of Munich. The course targets first-year graduate students and senior undergraduates with prior exposure to robotics. VNAV has the goal of preparing the students to perform research in robotics and vision-based navigation, with emphasis on drones and self-driving cars. The lectures cover a broad set of theoretical topics, including geometric control and trajectory optimization, 2D and 3D computer vision, visual and visual-inertial odometry, place recognition, simultaneous localization and mapping, and geometric deep learning for perception. The labs provide hands-on activities to ground the theoretical concepts and expose the students to advanced robotics platforms, simulators, and open-source software libraries. The course culminates in an open-ended final project. The course material is now publicly available on MIT OpenCourseWare and at *vnav.mit.edu/*.

## ACKNOWLEDGMENT

The authors gratefully acknowledge Valerio Varricchio, Mubarik Mohamoud, Antoni Rosinol, Kevin Doherty, Heng Yang, Yun Chang, Stewart Jamieson, Ziqi Lu, and Michael Everett for their contribution to creating and refining the VNAV material over the 4 years it has been taught at MIT.

This work was partially funded by Carlone's NSF CAREER award "Certifiable Perception for Autonomous Cyber-Physical Systems".

## REFERENCES

[1] S. Chan, J. Geng, M. Jong, and D. Lau, "Addressing the challenges in engineering classes: Harnessing active learning in a robotics course," in International Symposium on Educational Technology (ISET), 2018, pp. 162–164.

[2] S. Chen, A. Fishberg, E. Shimelis, J. Grimm, S. van Broekhoven, R. Shin, and S. Karaman, "A hands-on middle-school robotics software program at MIT," in IEEE Integrated STEM Education Conference (ISEC), 2020, pp. 1–8.

[3] S. Karaman, A. Anders, M. Boulet, J. Connor, K. Gregson, W. Guerra, O. Guldner, M. Mohamoud, B. Plancher, R. Shin, and J. Vivilecchia, "Project-based, collaborative, algorithmic robotics for high school students: Programming self-driving race cars at MIT," in IEEE Integrated STEM Education Conference (ISEC), 2017, pp. 195–203.

[4] R. Murphy, "Competing for a robotics education," IEEE Robotics Automation Magazine, vol. 8, no. 2, pp. 44–55, 2001.

[5] A. Baerveldt, T. Salomonsson, and B. B. Astrand, "Vision-guided mobile robots for design competitions," IEEE Robotics Automation Magazine, vol. 10, no. 2, pp. 38–44, 2003.

[6] V. Kirandziska and N. Ackovska, "Robotics course — a challenge for computer science students," in IEEE Global Engineering Education Conference (EDUCON), 2017, pp. 951–954.

[7] C. Berry, "Robotics education online flipping a traditional mobile robotics classroom," in IEEE Frontiers in Education Conference (FIE), 2017, pp. 1–6.

[8] Z. Roth, "The role of robotics in freshmen engineering curricula," in Proceedings of the 5th Biannual World Automation Congress, vol. 14, 2002, pp. 389–394.

[9] J. Garner, W. Smart, K. Bennett, D. Bruemmer, D. Few, and C. Roman, "The remote exploration program: a collaborative outreach approach to robotics education," in IEEE Intl. Conf. on Robotics and Automation (ICRA), vol. 2, 2004, pp. 1826–1830.

[10] P. Salas, "Teaching robotics to undergraduate computer science students: A different approach," in IEEE Frontiers in Education Conference (FIE), 2019, pp. 1–7.

[11] S. Jung, "Experiences in developing an experimental robotics course program for undergraduate education," IEEE Trans. on Education, vol. 56, no. 1, pp. 129–136, 2013.
[12] F. Gonzalez and J. Zalewski, "A robotic arm simulator software tool for use in introductory robotics courses," in IEEE Global Engineering Education Conference (EDUCON), 2014, pp. 861–866.
[13] S. Bruder and K. Wedeward, "An interactive online robotics course," in Proceedings World Automation Congress, vol. 15, 2004, pp. 27–32.
[14] B. Maxwell and L. Meeden, "Integrating robotics research with undergraduate education," IEEE Intelligent Systems and their Applications, vol. 15, no. 6, pp. 22–27, 2000.
[15] C. Nitschke, Y. Minami, M. Hiromoto, H. Ohshima, and T. Sato, "A quadrocopter automatic control contest as an example of interdisciplinary design education," in Intl. Conf. on Control, Automation and Systems (ICCAS), 2014, pp. 678–685.
[16] L. Paull, J. Tani, H. Ahn, J. Alonso-Mora, L. Carlone, M. Cap, Y. Chen, C. Choi, J. Dusek, Y. Fang, D. Hoehener, S.-Y. Liu, M. Novitzky, I. Okuyama, J. Pazis, G. Rosman, V. Varricchio, H.-C. Wang, D. Yershov, H. Zhao, M. Benjamin, C. Carr, M. Zuber, S. Karaman, E. Frazzoli, D. D. Vecchio, D. Rus, J. How, J. Leonard, and A. Censi, "Duckietown: an open, inexpensive and flexible and capable platform for autonomy education and research," in IEEE Intl. Conf. on Robotics and Automation (ICRA), 2017, pp. 1497–1504, (pdf).
[17] I. Brand, J. Roy, A. Ray, J. Oberlin, and S. Oberlix, "PiDrone: An autonomous educational drone using raspberry Pi and Python," in IEEE/RSJ Intl. Conf. on Intelligent Robots and Systems (IROS), 2018, pp. 1–7.
[18] L. Eller, T. Guerin, B. Huang, G. Warren, S. Yang, J. Roy, and S. Tellex, "Advanced autonomy on a low-cost educational drone platform," in IEEE/RSJ Intl. Conf. on Intelligent Robots and Systems (IROS), 2019, pp. 1032–1039.
[19] D. Rus, M. Vona, K. Quigley, "Eye-in-hand visual servoing curriculum for young students," IEEE Robotics Automation Magazine, vol. 17, no. 1, pp. 116–117, 2010.
[20] S. Thrun, "Artificial intelligence for robotics", Udacity, www.udacity.com/course/artificial-intelligence-for-robotics--cs373, 2021.
[21] K. Daniilidis and J. Shi, "Robotics: Perception, www.coursera.org/learn/robotics-perception," in Coursera, 2021, pp. 1032–1039.
[22] S. Waslander and J. Kelly, "Self-driving cars specialization, www.coursera.org/specializations/self-driving-cars," in Coursera, 2021, pp. 1032–1039.
[23] J. Shi, K. Daniilidis, and D. Lee, "Robotics: Vision intelligence and machine learning," in edX, 2021.
[24] J. Tani, A. Censi, and E. Frazzoli, "Self-driving cars with Duckietown," in edX, www.edx.org/course/self-driving-cars-with-duckietown, 2021.
[25] A. Agnihotri, M. O'Kelly, R. Mangharam, and H. Abbas, "Teaching autonomous systems at 1/10th-scale, f1tenth.org/about.html," 2020.
[26] MIT Lincoln Laboratory, "Beaver Works Summer Institute," beaverworks.ll.mit.edu/CMS/bw/bwsi, 2019.
[27] T. Barfoot, State Estimation for Robotics. Cambridge University Press, 2017.
[28] L. Carlone et al., "Visual navigation for autonomous vehicles," in MIT OpenCourseWare, ocw.mit.edu/courses/aeronautics-and-astronautics/16-485-visual-navigation-for-autonomous-vehicles-vnav-fall-2020/index.htm, 2021.
[29] S. Chacon and B. Straub, Pro git. Springer Nature, 2014.
[30] M. Quigley et al. "ROS: an open-source Robot Operating System." ICRA workshop on open-source software. Vol. 3. No. 3.2. 2009.
[31] G. Bradski, and A. Kaehler. Learning OpenCV: Computer vision with the OpenCV library. " O'Reilly Media, Inc.", 2008.
[32] F. Dellaert. Factor graphs and GTSAM: A hands-on introduction. Georgia Institute of Technology, 2012.
[33] R. Mur-Artal et al. "ORB-SLAM: a versatile and accurate monocular SLAM system." IEEE transactions on robotics 31.5 (2015): 1147-1163.
[34] R. Mur-Artal, and J. D. Tardós. "ORB-SLAM2: An open-source slam system for monocular, stereo, and RGB-D cameras." IEEE transactions on robotics 33.5 (2017): 1255-1262.
[35] C. Campos et al. "ORB-SLAM3: An Accurate Open-Source Library for Visual, Visual–Inertial, and Multimap SLAM." IEEE Transactions on Robotics (2021).
[36] D. Gálvez-López, Dorian, and J. D. Tardos. "Bags of binary words for fast place recognition in image sequences." IEEE Transactions on Robotics 28.5 (2012): 1188-1197.
[37] Papers with code, "ML Reproducibility Challenge 2021 Edition." paperswithcode.com/rc2021, 2021.
[38] E. Crawley et al. "The CDIO approach." Rethinking engineering education. Springer, Cham, 2014. 11-45.
[39] Y. Ma et al. "An invitation to 3-d vision: from images to geometric models." Vol. 26. New York: springer, 2004.
[40] F. Dellaert and M. Kaess. "Factor graphs for robot perception." Foundations and Trends in Robotics 6.1-2 (2017): 1-139.
[41] Intel, "Overview of The Intel Aero Ready To Fly Drone." www.intel.com/content/www/us/en/support/articles/000023271/drones/development-drones.html. 2022.
[42] L. Meier, D. Honegger, and M. Pollefeys. "PX4: A node-based multithreaded open-source robotics framework for deeply embedded platforms." IEEE international conference on robotics and automation (ICRA), 2015.
[43] Defense Advanced Research Projects Agency, "DARPA Subterranean Challenge", www.subtchallenge.com/, 2021.
[44] Nvidia, "The World'S Smallest AI Supercomputer", www.nvidia.com/en-us/autonomous-machines/embedded-systems/jetson-xavier-nx/, 2022.
[45] L. Kneip and P. Furgale. "OpenGV: A unified and generalized approach to real-time calibrated geometric vision." 2014 IEEE International Conference on Robotics and Automation (ICRA). IEEE, 2014.
[46] M. Achtelik et al., "mav_trajectory_generation: Polynomial trajectory generation and optimization, especially for rotary-wing MAVs", github.com/ethz-asl/mav_trajectory_generation, 2016.
[47] N. Roy, A. Schoellig, S. Thrun, R. D'Andrea, S. Lupashin, J. Lussier, A. Brown, "Flying car and autonomous flight engineering", Udacity, www.udacity.com/course/flying-car-nanodegree--nd787, 2022.
[48] C. Stachniss, "Online Lectures", www.ipb.uni-bonn.de/teaching/, 2021.
[49] Docker's developers, "Dockers", www.docker.com/, 2021.
[50] M. Hohenwarter et al., "GeoGebra 4.4", www.geogebra.org, 2013.
[51] R. Kuemmerle, G. Grisetti, H. Strasdat, K. Konolige, and W. Burgard, "g2o: A General Framework for Graph Optimization", IEEE International Conference on Robotics and Automation (ICRA), 2011.
[52] Manim's developers, "Manim – Mathematical Animation Engine" github.com/3b1b/manim, 2020.

## AUTHOR INFORMATION

**Luca Carlone,** Associate Professor, Dept. of Aeronautics and Astronautics, Massachusetts Institute of Technology.

**Kasra Khosoussi**, Senior Research Scientist, Data61, CSIRO.

**Vasileios Tzoumas**, Assistant Professor, Dept. of Aerospace Engineering, University of Michigan.

**Golnaz Habibi**, Assistant Professor, School of Computer Science, University of Oklahoma.

**Markus Ryll**, Assistant Professor, Dept. of Aerospace and Geodesy, Technical University of Munich.

**Rajat Talak**, Postdoctoral Associate, Dept. of Aeronautics and Astronautics, Massachusetts Institute of Technology.

**Jingnan Shi**, PhD Student, Dept. of Aeronautics and Astronautics, Massachusetts Institute of Technology.

**Pasquale Antonante**, PhD Student, Dept. of Aeronautics and Astronautics, Massachusetts Institute of Technology.